\documentclass[11pt]{article}
\usepackage{acl2015}
\usepackage{times}
\usepackage{url}
\usepackage{latexsym}
\usepackage{tikz}
\usepackage{graphicx}
\usepackage{amsmath}
\usepackage{mathtools}
\usepackage{amssymb}
\usepackage{subcaption}
\DeclareMathOperator*{\argmax}{arg\,max}

\title{A Unified Tagging Solution:\\Bidirectional LSTM Recurrent Neural Network with Word Embedding}

\author{Peilu Wang$^{1,2}$\thanks{*Work performed as an intern in the Speech Group, Microsoft Research Asia}, Yao Qian$^3$, Frank K. Soong$^2$, Lei He$^2$, Hai Zhao$^1$\\
  $^1$Shanghai Jiao Tong University, Shanghai, China\\
  $^2$Microsoft Research Asian, Beijing, China \\
  $^3$Educational Testing Service Research, USA\\
  {\tt \{v-peiwan, frankkps, helei\}@microsoft.com,}\\
  {\tt zhaohai@cs.sjtu.edu.cn, yqian@ets.org} }

\date{}

\begin{document}
\maketitle
\begin{abstract}
Bidirectional Long Short-Term Memory Recurrent Neural Network (BLSTM-RNN) has been shown to be very effective for modeling and predicting sequential data, e.g. speech utterances or handwritten documents.
In this study, we propose to use BLSTM-RNN for a unified tagging solution that can be applied to various tagging tasks including part-of-speech tagging, chunking and named entity recognition.
Instead of exploiting specific features carefully optimized for each task, our solution only uses one set of task-independent features and internal representations learnt from unlabeled text for all tasks.
Requiring no task specific knowledge or sophisticated feature engineering, our approach gets nearly state-of-the-art performance in all these three tagging tasks.
\end{abstract}

\section{Introduction}


%
%

Long short-term memory (LSTM) \cite{1997_Sepp_NC_LongShort} is a type of promising recurrent architecture, able to bridge long time lags between relevant input and target output, and thereby incorporate long range context.
This type of structure is theoretically well suited and has been proven a powerful model for tagging tasks.
For applications in natural language processing (NLP), LSTM has proved advantageous in language modeling \cite{2012_Martin_INTERSPEECH_LSTMNeural,2015_Martin_ASLP_FromFeedforward}, language understanding \cite{2013_Kaisheng_INTERSPEECH_RecurrentNeural,2013_Mesnil_INTERSPEECH_InvestigationOf}, and machine translation \cite{2014_Martin_EMNLP_TranslationModeling}.
A bidirectional LSTM (BLSTM) \cite{1997_Mike_SP_BidirectionalRecurrent}, furthermore, introduces two independent layers to accumulate contextual information from the past and future histories.
It seems natural to expect BLSTM to be an effective model for tagging tasks in NLP while to our best knowledge no successful case of this application has been reported.

In this work, we apply BLSTM-RNN to three typical tagging tasks: part-of-speech (POS) tagging, chunking and named entity recognition (NER).
As a neural network model, BLSTM-RNN is awkward for using discrete features.
Since these features have to be represented as one-hot vector usually with very large size, using this type of features would lead to too large input layer to operate.
Therefore, we only use word form and simple capital features, disregarding of all the other discrete conventional NLP features, such as morphological features.
Using such simple task independent features assures our model quite unified so that it can be directly applied to various tagging tasks.

To further improve the performance of our approach without disrupting the universality, we introduce word embedding, which is a real-valued vector associated with each word.
It is an internal representation that is considered containing syntactic and semantic information and has shown a very attractive feature for various NLP tasks \cite{2008_Collobert_ICML_AUnified,2010_Joseph_ACL_WordRepresentations,2011_Ronan_JMLR_NaturalLanguage}. 
Word embedding can be obtained by training a neural network language model \cite{2006_Bengio_IML_NeuralProbabilistic}, shallow neural network \cite{2013_Mikolov_ARXIV_EfficientEstimation,2014_Jeffrey_EMNLP_GloveGlobal,2011_Ronan_JMLR_NaturalLanguage}, or a recurrent neural network \cite{2010_Mikolov_INTERSPEECH_RecurrenntNeural}.
In this work, we also propose a novel method to train word embedding on unlabeled data with BLSTM-RNN.

The main contributions of this work include: 
First, it shows an effective way to use BLSTM-RNN for dealing with various NLP tagging tasks.
Second, it proposes a unified tagging system that can get competitive tagging accuracy without using any task specific features, which makes this approach more practical for tagging tasks that lack of prior knowledge.

The remainder of this paper is organized as follows.
Section 2 gives a brief introduction of BLSTM architecture.
Section 3 describes the BLSTM-RNN based tagging approach and Section 4 introduces the training and usage of word embeddings.
Section 5 presents experimental results.
Section 6 discusses related works and concluding remarks are given in Section 7.

\section{Bidirectional LSTM Architecture}
Recurrent neural network (RNN) is a kind of artificial neural network that contains cyclic connections, which can model contextual information dynamically.
Given an input sequence $x_1,x_2,...,x_n$, a standard RNN computes the output vector $y_t$ of each word $x_t$ by iterating the following equations from $t$ = 1 to $n$:
\begin{equation*}
\begin{split}
h_t&=H(W_{xh}x_t+W_{hh}h_{t-1}+b_{h})\\
y_t&=W_{hy}h_{t}+b_y
\end{split}
\end{equation*}
where $h_t$ is the vector of hidden states, $W$ denotes weight matrix connecting two layers (e.g. $W_{xh}$ is the weights between input and hidden layer), $b$ denotes bias vector (e.g. $b_h$ is the bias vector of hidden layer) and $H$ is the activation function of hidden layer.
Note that $h_t$ persists information from previous step's hidden state $h_{t-1}$, and thus theoretically RNN can make use of all input history.

However, in practice, the range of input history that can be accessed is limited, since the influence 
\begin{figure}[h]
\small
\centering
\includegraphics[width=3in]{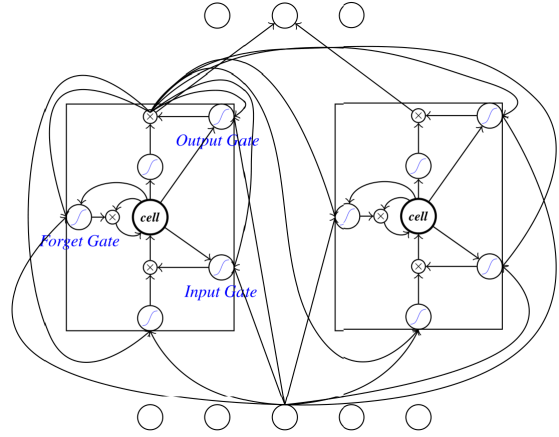}
\caption{An LSTM network. The network has five input units, a hidden layer composed of two LSTM memory blocks and three output units. Each memory block has four inputs but only one output. 
All connections of the left block are drawn and other connections are skipped for simplicity.
}\label{lstmblock}
\end{figure}
of a given input would decay or blow up exponentially as it circulates around the hidden states, which is known as vanishing gradient problem \cite{2001_Hochreiter_IEEE_GradientFlow}.
The most effective solution of this problem so far is the long short-term memory (LSTM) architecture \cite{1997_Sepp_NC_LongShort}.
An LSTM network is formed like the standard RNN except that the self-connected hidden units are replaced by special designed units called memory blocks as illustrated in Figure \ref{lstmblock}.
The output of LSTM hidden layer $h_t$ given input $x_t$ is computed as following composite function \cite{2013_Alex_ICASSP_SpeechRecognition}:
\begin{equation*}
\begin{split}
i_t &= \sigma (W_{xi}x_t+W_{hi}h_{t-1}+W_{ci}c_{t-1}+b_i) \\
f_t &= \sigma (W_{xf}x_t+W_{hf}h_{t-1}+W_{cf}c_{t-1}+b_f) \\
c_t &= f_{t}c_{t-1}+i_{t}\tanh(W_{xc}x_t+W_{hc}h_{t-1}+b_c) \\
o_t &= \sigma (W_{xo}x_{t}+W_{ho}h_{t-1}+W_{co}c_t+b_o) \\
h_t &= o_t \tanh(c_t)
\end{split}
\end{equation*}
where $\sigma$ is the logistic sigmoid function, and $i$, $f$, $o$ and $c$ are respectively the \textit{input gate}, \textit{forget gate}, \textit{output gate} and \textit{cell} activation vectors.
Weights matrices are represented as arrows in Figure \ref{lstmblock}.
These multiple gates allow the cell in LSTM memory block to store information over long periods of time, thereby avoiding the vanishing gradient problem.
More interpretation about this architecture can be found in \cite{2012_Alex_BOOK_SupervisedSequence}.

Another shortcoming of conventional RNN is that only historic context can be exploited. 
In a typical tagging task where the whole sentence is given, it is helpful to exploit future context as well.
\begin{figure}[h]
\small
\centering
\tikzset{
label/.style={
  rectangle,
  draw=none,
  text centered,
  inner sep=0.6em,
},
namelabel/.style={
  rectangle,
  draw=none,
  text width=12em,
  inner sep=0.6em,
  align=left
},
bigcircle/.style={
  rectangle,
  draw,
  minimum width=3em,
  text centered,
  inner sep=0.3em,
}
}

\scriptsize
\begin{tikzpicture}[x=1em, y=1em, >=stealth]

\node[bigcircle](bht1) at (-5.5,0){$\overleftarrow{h}_{t-1}$};
\node[bigcircle](bht2) at (0,0){$\overleftarrow{h}_{t}$};
\node[bigcircle](bht3) at (5.5,0){$\overleftarrow{h}_{t+1}$};

\node[bigcircle](fht1) at (-5.5,-4){$\overrightarrow{h}_{t-1}$};
\node[bigcircle](fht2) at (0,-4){$\overrightarrow{h}_{t}$};
\node[bigcircle](fht3) at (5.5,-4){$\overrightarrow{h}_{t+1}$};

\node[label](x0) at (-9,-8){$\dots$};
\node[label](x1) at (-5.5,-8){$x_{t-1}$};
\node[label](x2) at (0,-8){$x_{t}$};
\node[label](x3) at (5.5,-8){$x_{t+1}$};
\node[label](x4) at (9,-8){$\dots$};

\node[label](y0) at (-9,4){$\dots$};
\node[label](y1) at (-5.5,4){$y_{t-1}$};
\node[label](y2) at (0,4){$y_{t}$};
\node[label](y3) at (5.5,4){$y_{t+1}$};
\node[label](y4) at (9,4){$\dots$};

\node[namelabel](textoutputs) at (-12,4){Outputs};
\node[namelabel](textbklayer) at (-12,0){Backward Layer};
\node[namelabel](textflayer) at (-12,-4){Forward Layer};
\node[namelabel](textinputs) at (-12,-8){Inputs};

\draw[->](x1) to[bend left=50] (bht1);
\draw[->](x2) to[bend left=50] (bht2);
\draw[->](x3) to[bend left=50] (bht3);

\draw[->](fht1) to[bend right=50] (y1);
\draw[->](fht2) to[bend right=50] (y2);
\draw[->](fht3) to[bend right=50] (y3);

\draw[->] (-10,-4) -- (fht1);
\draw[->](fht1) -- (fht2);
\draw[->](fht2) -- (fht3);
\draw[->](fht3) -- (10,-4);

\draw[->] (10,0) -- (bht3);
\draw[->] (bht3) -- (bht2);
\draw[->] (bht2) -- (bht1);
\draw[->] (bht1) -- (-10,0);

\draw[->] (x1) -- (fht1);
\draw[->] (x2) -- (fht2);
\draw[->] (x3) -- (fht3);

\draw[->] (bht1) -- (y1);
\draw[->] (bht2) -- (y2);
\draw[->] (bht3) -- (y3);

\end{tikzpicture} 
\caption{Bidirectional RNN}\label{brnn}
\end{figure}
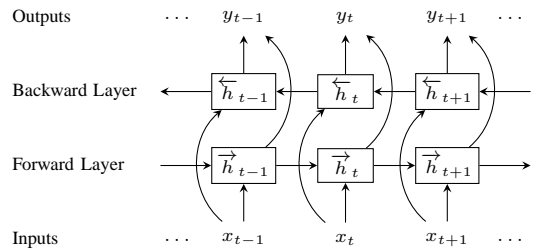
Bidirectional RNN (BRNN) \cite{1997_Mike_SP_BidirectionalRecurrent} offers an effective solution that can access both the preceding and succeeding contexts by involving two separate hidden layers.
As illustrated in Figure \ref{brnn}, BRNN first computes the forward hidden sequence $\overrightarrow{h}$ and the backward hidden sequence $\overleftarrow{h}$ respectively, and then combines $\overrightarrow{h_t}$ and $\overleftarrow{h_t}$ to generate output $y_t$.
The process can be expressed as:
\begin{equation*}
\begin{split}
\overrightarrow{h_t} &= H(W_{x\overrightarrow{h}}x_t+W_{\overrightarrow{h}\overrightarrow{h}}\overrightarrow{h}_{t-1}+b_{\overrightarrow{h}}) \\
\overleftarrow{h_t} &= H(W_{x\overleftarrow{h}}x_t+W_{\overleftarrow{h}\overleftarrow{h}}\overleftarrow{h}_{t+1}+b_{\overleftarrow{h}}) \\
y_t &= W_{\overrightarrow{h}y}\overrightarrow{h}_t+W_{\overleftarrow{h}y}\overleftarrow{h}_t+b_y
\end{split}
\end{equation*}

Replacing the hidden states in BRNN with LSTM memory blocks gives bidirectional LSTM (BLSTM) \cite{2005_Graves_NN_FramewisePhoneme}, i.e., the main architecture used in this paper, which can incorporate long periods of contextual information from both directions.
Besides, as feed-forward layers stacked in deep neural networks, the BLSTM layer can also be stacked on the top of the others to form a deep BLSTM architecture.
As a type of RNN, deep BLSTM can be trained via various gradient-based algorithms designed for general RNN, for example, real-time recurrent learning (RTRL) and back-propagation through time (BPTT).
In this work, we use the BPTT algorithm as described in \cite{2012_Alex_BOOK_SupervisedSequence} since it is conceptually simple and efficient for computation.

\section{Tagging System}
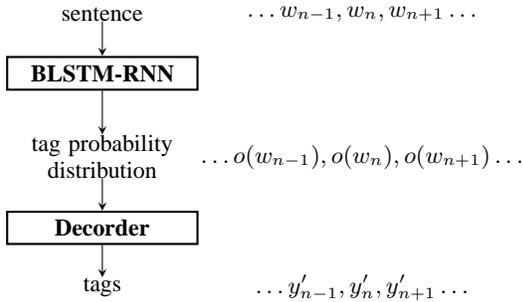
\begin{figure}[h]
\small
\centering
\tikzset{
label/.style={
  rectangle,
  draw=none,
  text centered,
  inner sep=0em,
},
}

\begin{tikzpicture}[x=1em, y=1em, >=stealth]
\node[label](input) at (0,0) {sentence};
\node[label](inputsample) at (11,0) {$\dots w_{n-1}, w_{n}, w_{n+1} \dots$};
\node[rectangle,draw, minimum width=8em, thick](BLSTM) at (0,-2.5){\textbf{BLSTM-RNN}};
\node[label, text width=8em](ioutput) at (0,-6) {tag probability distribution};
\node[label](ioutputsample) at (11,-6) {$\dots o(w_{n-1}), o(w_{n}), o(w_{n+1}) \dots$};
\node[rectangle,draw, minimum width=8em, thick](Decorder) at (0,-9){\textbf{Decorder}};
\node[label, text width=8em](output) at (0,-11.5) {tags};
\node[label](inputsample) at (11,-11.5) {$\dots y'_{n-1}, y'_{n}, y'_{n+1} \dots$};

\draw[->](input) -- (BLSTM);
\draw[->](BLSTM) -- (ioutput);
\draw[->](ioutput) -- (Decorder);
\draw[->](Decorder) -- (output);

\end{tikzpicture} 
\caption{BLSTM-RNN based tagging system}\label{sysflow}
\end{figure}

The schematic diagram of BLSTM-RNN based tagging system is illustrated in Figure \ref{sysflow}.
 Given a sentence $w_1,w_2,...,w_n$ with tags $y_1,y_2,...,y_n$, BLSTM-RNN is first used to predict the tag probability distribution $o(w_i)$ of each word,
 then a decoding algorithm is proposed to generate the final predicted tags $y'_1,y'_2,...,y'_n$.
 \subsection{BLSTM-RNN for tagging}
\begin{figure}[h]
\small
\centering
\definecolor{royalblue}{RGB}{67,107,149}
\tikzset{
wordvector/.style={
  rectangle,
  draw,
  text width=12em,
  text centered,
  minimum height=1.2em,
  inner sep=0em
},
input/.style={
  rectangle,
  draw,
  text width=6em,
  text centered,
  minimum height=1.2em,
  inner sep=0em
},
output/.style={
  rectangle,
  draw,
  text width=6em,
  text centered,
  minimum height=1.2em,
  inner sep=0em
},
wordembedding/.style={
  rectangle,
  draw,
  text width=6em,
  text centered,
  minimum height=1.2em,
  inner sep=0em
},
NN/.style={
  rectangle,
  draw,
  text width=8em,
  text centered,
  minimum height=1.2em,
  fill=black!20,
},
block_noborder/.style={
  rectangle,
  draw=none,
  text width=8em,
  text centered,
  minimum height=1em,
  inner sep=0em,
  font={\itshape}
},
weightsymbol/.style={
  rectangle,
  draw=none,
  text width=2em,
  text centered,
  minimum height=1em,
  inner sep=0em,
},
}

\begin{tikzpicture}[x=1em, y=1em, >=stealth]

\node[wordvector] (w1) at (0,0) {$\underline{w_{i}}$};
\node[input] (w2) at (10,0){$f(w_{i})$};

\node[wordembedding] (we1) at (4,3) {$I_i$};
\node[block_noborder](tinput) at (11.5,3) {input layer};

\node[NN] (NN) at (4,6) {BLSTM};
\node[block_noborder](tinput) at (11.5,6) {hidden layer};

\node[output] (w3) at (4,9){$o(w_i)$};
\node[block_noborder](tinput) at (11.5,9) {output layer};

\draw [-,dotted,thick] (-6,0.6) -- (1,2.4);
\draw [-,dotted,thick] (6,0.6) -- (7,2.4);
\node[weightsymbol](w1) at (1,1.3) {$W_1$};

\draw [-,dotted,thick] (7,0.6) -- (1,2.4);
\draw [-,dotted,thick] (13,0.6) -- (7,2.4);
\node[weightsymbol](w2) at (8,1.3) {$W_2$};

\draw [-,dotted,thick] (1,3.6) -- (-0.4,5.4);
\draw [-,dotted,thick] (7,3.6) -- (8.4,5.4);

\draw [-,dotted,thick] (-0.4,6.6) -- (1,8.4);
\draw [-,dotted,thick] (8.4,6.6) -- (7,8.4);

\end{tikzpicture}
\caption{Usage of BLSTM-RNN for tagging}\label{rnntagging}
\end{figure}
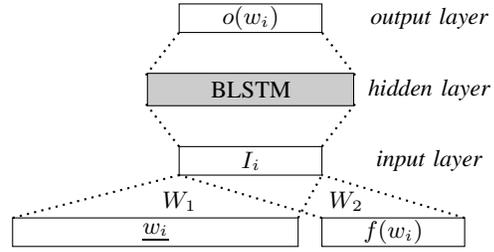

The usage of BLSTM RNN is illustrated in Figure \ref{rnntagging}.
Here $\underline{w_i}$ is the one hot representation of the current word which is a binary vector with dimension $|V|$ where $V$ is the vocabulary.
To reduce $|V|$, each letter of input word is transferred to its lowercase.
The upper case information is kept by introducing a three-dimensional binary vector $f(w_i)$ to indicate if $w_i$ is full lowercase, full uppercase or leading with a capital letter.
The input vector $I_i$ of the network is computed as:
\begin{equation*}
I_i=W_1\underline{w_i}+W_2f(w_i)
\end{equation*}
where $W_1$ and $W_2$ are weight matrixes connecting two layers.
$W_1\underline{w_i}$ is also known as the word embedding of $w_i$ which is a real-valued vector with a much smaller dimension than $\underline{w_i}$.
In practice, to reduce the computational cost, $W_1$ is implemented as a lookup table, $W_1\underline{w_i}$ is returned by referring to $w_i$'s word embedding stored in this table.
The output layer is a softmax layer whose dimension is the number of tag types.
It outputs the tag probability distribution of word $w_i$.


\subsection{Decoding\label{secdecode}}
\begin{figure}[h]
\small
\centering
\tikzset{
circlenode/.style={
  circle,
  draw,
  minimum size=0.3cm
},
circlemissing/.style={
  draw=none, 
  scale=2,
  text height=0.1cm,
  execute at begin node=\color{black}$\vdots$
},
circlehmissing/.style={
  draw=none, 
  scale=2,
  text height=0.1cm,
  execute at begin node=\color{black}$\dots$
},
}

\begin{tikzpicture}[x=1cm, y=0.8cm, >=stealth]
\node [align=center] (output0_1) at (-1,-0.8) {$t_1$};
\node [align=center] (output0_2) at (-1,-1.6) {$t_2$};
\node [align=center] (output0_n) at (-1,-3.2) {$t_m$};
\node [align=center] at (0,0) {$o(w_1)$};
\node [circlenode,red,thick] (output1_1) at (0,-0.8) {};
\node [circlenode] (output1_2) at (0,-1.6) {};
\node [circlemissing] (output1_m) at (0,-2.6) {};
\node [circlenode] (output1_n) at (0,-3.2) {};
\node [align=center] at (1.2,0) {$o(w_2)$};
\node [circlenode] (output2_1) at (1.2,-0.8) {};
\node [circlenode,red,thick] (output2_2) at (1.2,-1.6) {};
\node [circlemissing] (output2_m) at (1.2,-2.6) {};
\node [circlenode] (output2_n) at (1.2,-3.2) {};
\node [align=center] at (2.4,0) {$o(w_2)$};
\node [circlenode] (output3_1) at (2.4,-0.8) {};
\node [circlenode] (output3_2) at (2.4,-1.6) {};
\node [circlemissing] (output3_m) at (2.4,-2.6) {};
\node [circlenode,red,thick] (output3_n) at (2.4,-3.2) {};
\node [circlehmissing] at (3.6,0) {};
\node [circlehmissing] at (3.6,-1.6) {};
\node [align=center] at (4.8,0) {$o(w_n)$};
\node [circlenode] (output4_1) at (4.8,-0.8) {};
\node [circlenode,red,thick] (output4_2) at (4.8,-1.6) {};
\node [circlemissing] (output4_m) at (4.8,-2.6) {};
\node [circlenode] (output4_n) at (4.8,-3.2) {};
\draw [->, blue] (output1_1) -- (output2_1);
\draw [->, red, thick] (output1_1) -- (output2_2);
\draw [->, blue] (output1_1) -- (output2_n);

\draw [->, blue] (output1_2) -- (output2_1);
\draw [->, blue] (output1_2) -- (output2_2);
\draw [->, blue] (output1_2) -- (output2_n);

\draw [->, blue] (output1_n) -- (output2_1);
\draw [->, blue] (output1_n) -- (output2_2);
\draw [->, blue] (output1_n) -- (output2_n);
\draw [->, blue] (output2_1) -- (output3_1);
\draw [->, blue] (output2_1) -- (output3_2);
\draw [->, blue] (output2_1) -- (output3_n);

\draw [->, blue] (output2_2) -- (output3_1);
\draw [->, blue] (output2_2) -- (output3_2);
\draw [->, red,thick] (output2_2) -- (output3_n);

\draw [->, blue] (output2_n) -- (output3_1);
\draw [->, blue] (output2_n) -- (output3_2);
\draw [->, blue] (output2_n) -- (output3_n);

\node [align=center, blue] at (0.5,-3.6) {$A$};
\node [align=center, blue] at (1.8,-3.6) {$A$};
\end{tikzpicture}
\caption{Illustration of decoding. $o(w_i)$ is the output of BLSTM-RNN, a probability distribution of $m$ tag types. }\label{decoding}
\end{figure}
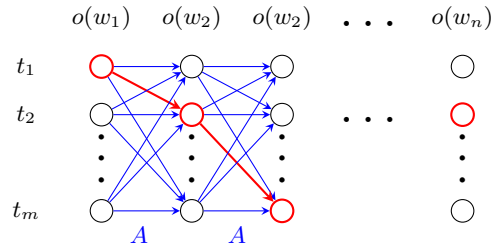

According to BLSTM-RNN, the obtained probability distribution of each step is supposed independent with each other.
However, in some tasks such as NER and chunking, tags are highly related with each other and a few of types of tags can only follow specific types of tags.
To make use of this kind of labeling constraints, we introduce a transition matrix $A$ between each step's output as illustrated in Figure \ref{decoding}.
Each circle represents a tag probability predicted by BLSTM-RNN, $A_{ij}$ stores the score of transition from tag $t_i$ to $t_j$.
The score is determined in a very simple way that if $t_j$ appears immediately behind $t_i$ in training corpus, $A_{ij}$ is 1, otherwise 0.
It implies that tag bigrams that do not appear in training corpus are supposed invalid and would not appear in test case, no matter whether they are actually valid.
The score of a sentence $w_1,w_2,...,w_n$ ($[w]_1^n$ for short) along a path of tags $y_1,y_2,...,y_n$ ($[y]_1^n$ for short) is then given by the product of transition score and BLSTM-RNN output probability:
\begin{equation*}
s([w]_1^n, [y]_1^n)=\prod_{i=1}^n (A_{y_{i-1}y_{i}}\times o(w_i)_{y{i}})
\end{equation*}
The goal of decoding is to find the path which gives the highest sentence score:
\begin{equation*}
[y']_1^n=\argmax \limits_{[y]_1^n} s([w]_1^n, [y]_1^n)
\end{equation*}
This is a typical dynamic programming problem and can be solved with Viterbi algorithm \cite{1967_Viberbi_IT_ErrorBounds}.


\section{Word Embedding\label{secwe}}

As a neural network, BLSTM-RNN can easily adopt already trained word embedding by initializing $W_1$ (illustrated in Figure \ref{rnntagging}) with those external embeddings.
Currently, many word embeddings trained on very large corpora are available on line.
However, these embeddings are trained by neural networks that are very different from BLSTM-RNN.
This inconsistency is supposed as an shortcoming to make the most of these trained word embeddings.
To conquer this shortcoming, we also propose a novel method to train word embedding on unlabeled data with BLSTM-RNN.

In this method, BLSTM-RNN is applied to perform a tagging task with only two types of tags to predict: incorrect/correct.
The input is a sequence of words which is a normal sentence with some words replaced by words randomly chosen from vocabulary.
The words to be replaced are chosen randomly from the sentence.
For those replaced words, their tags are 0 (incorrect) and for those that are not replaced, their tags are 1 (correct).
A simple sample is shown in Figure \ref{sensample}.
\begin{figure}[h]
\centering
\small
\tikzset{
label/.style={
  rectangle,
  draw=none,
  text centered,
  inner sep=0em,
},
labelword/.style={
  rectangle,
  text width=3em,
  align=left,
  inner sep=0em,
},
}

\begin{tikzpicture}[x=1em, y=1em, >=stealth]
\node[label,text width=10em, align=left](t1) at (0,0) {\textbf{original sentence:}};
\node[labelword](w1) at (-3.5,-1) {\textit{They}};
\node[labelword](w2) at (-3.5+3,-1) {\textit{seem}};
\node[labelword](w3) at (-3.5+3+3,-1) {\textit{to}};
\node[labelword](w4) at (-3.5+3+3+2,-1) {\textit{be}};
\node[labelword](w5) at (-3.5+3+3+2+2,-1) {\textit{prepared}};
\node[labelword](w6) at (-3.5+3+3+2+2+4.5,-1) {\textit{to}};
\node[labelword](w7) at (-3.5+3+3+2+2+4.5+2,-1) {\textit{make}};
\node[labelword](w7) at (-3.5+3+3+2+2+4.5+2+3,-1) {$\dots$};

\node[label,text width=10em, align=left](t2) at (0,-3) {\textbf{input sentence:}};
\node[labelword](w1) at (-3.5,-4) {\textit{They}};
\node[labelword](w2) at (-3.5+3,-4) {\textit{\underline{beast}}};
\node[labelword](w3) at (-3.5+3+3,-4) {\textit{to}};
\node[labelword](w4) at (-3.5+3+3+2,-4) {\textit{be}};
\node[labelword](w5) at (-3.5+3+3+2+2,-4) {\textit{\underline{austere}}};
\node[labelword](w6) at (-3.5+3+3+2+2+4.5,-4) {\textit{to}};
\node[labelword](w7) at (-3.5+3+3+2+2+4.5+2,-4) {\textit{make}};
\node[labelword](w7) at (-3.5+3+3+2+2+4.5+2+3,-4) {$\dots$};

\node[label,text width=10em, align=left](t3) at (0,-6) {\textbf{tags:}};
\node[labelword](w1) at (-3.5,-7) {1};
\node[labelword](w2) at (-3.5+3,-7) {0};
\node[labelword](w3) at (-3.5+3+3,-7) {1};
\node[labelword](w4) at (-3.5+3+3+2,-7) {1};
\node[labelword](w5) at (-3.5+3+3+2+2,-7) {0};
\node[labelword](w6) at (-3.5+3+3+2+2+4.5,-7) {1};
\node[labelword](w7) at (-3.5+3+3+2+2+4.5+2,-7) {1};
\node[labelword](w7) at (-3.5+3+3+2+2+4.5+2+3,-7) {$\dots$};

\end{tikzpicture} 
\caption{Sample of constructed corpus for training word embedding.}\label{sensample}
\end{figure}
Although it is possible that some replaced words are also reasonable in the sentence, they are still considered ``incorrect''.
Then BLSTM-RNN is trained to minimize the binary classification error on the training corpus.
The neural network structure is the same as that in Figure \ref{rnntagging}.
When the neural network is trained, $W_1$ contains all trained word embeddings.

\section{Experiments}

All of our approaches are implemented based on CURRENT \cite{2014_Felix_JMLR_IntroducingCurrennt}, an open source GPU-based toolkit of BLSTM-RNN.
For constructing and training the neural network, we follow the default setup of CURRENT:
The activation functions of input layer and hidden layers are logistic function, while the output layer uses softmax function for multiclassification. 
Neural network is trained using statistical gradient descent algorithm with constant learning rate.

In all experiments, consecutive digits occurring within a word are relpaced with the symbol ``\#'' .
For example, both words ``Tel192'' and ``Tel6'' are converted into ``Tel\#''.
The vocabulary we used is the most common 100,000 words in North American news corpus \cite{2008_David_LDC_NorthAmerican}, plus one single ``UNK'' symbol for replacing all out of vocabulary words.

\subsection{Tasks}
In this section, we briefly introduce three typical tagging tasks and their experimental setup on which we evaluate the performance of the proposed approach: part-of-speech tagging (POS), chunking (CHUNK) and named entity recognition (NER).

\textbf{POS} is the task of labeling each word with its part of speech, e.g. noun, verb, adjective, etc.
Our POS tagging experiment is conducted on the Wall Street Journal data from Penn Treebank III \cite{1993_Mitchell_CL_BuildingaLarge}.
Training, development and test sets are split following setup in \cite{2002_Michael_ACL_DiscriminativeTraining}.
Table \ref{poscorpus} lists statistical information of the three data sets.
\begin{table}[h]
\centering
\small
\begin{tabular}{|c|c|c|c|}
\hline
\textbf{Data Set} & \textbf{WSJ sec. IDs} & \textbf{Sentences\#} & \textbf{Tokens\#} \\
\hline
Training & 0-18 & 38,219 & 912,344 \\
\hline
Develop & 19-21 & 5,527 & 131,768 \\
\hline
Test & 22-24 & 5,462 & 129,654 \\
\hline
\hline
\multicolumn{3}{|c|}{\# of tag types} & \multicolumn{1}{c|}{45} \\
\hline
\end{tabular}
\caption{POS tagging corpus (WSJ in PTB III)}\label{poscorpus}
\end{table}
Performance is evaluated by the accuracy of predicted tags on test set.

\textbf{CHUNK}, also known as shallow parsing, divides a sentence into phrases that each phrase contains syntactically related words, such as noun phrase (NP), verb phrase (VP), etc.
To identify the phrase boundaries, we use a commonly used IOBES tagging scheme that further fractionizes each tag type into four subtypes to indicate whether the word is inside (I), outside (O), begin (B), end(E) a multiple words chunk or a single word chunk (S).
We conduct our experiment on a standard experimental setup of CHUNK according to the CoNLL-2000 shared task \cite{2000_Erik_CONLL_IntroductionTo}.
Basic information about this setup is listed in Table \ref{chunkcorpus}.
\begin{table}[h]
\centering
\small
\begin{tabular}{|c|c|c|c|}
\hline
\textbf{Data Set} & \textbf{WSJ sec. IDs} & \textbf{Sentences\#} & \textbf{Tokens\#} \\
\hline
Training & 15-18 & 8,936 & 211,727 \\
\hline
Develop & N/A & N/A & N/A \\
\hline
Test & 20 & 2,012 & 47,377 \\
\hline
\hline
\multicolumn{3}{|c|}{\# of tag types (IOBES scheme)} & \multicolumn{1}{c|}{42} \\
\hline
\end{tabular}
\caption{CHUNK corpus (WSJ in PTB III: CoNLL-2000)}\label{chunkcorpus}
\end{table}
Performance is assessed by the F1 score computed by the evaluation script released by the CoNLL-2000 shared task\footnote{\url{http://www.cnts.ua.ac.be/conll2000/chunking}}.

\textbf{NER} recognizes phrases of named entities such as names of persons, organizations and locations.
IOBES tagging scheme is also applied in this task.
Our experimental setup follows the CoNLL-2003 shared task \cite{2003_Sang_CONLL_IntroductionTo}.
Table \ref{nercorpus} shows its basic information.
\begin{table}[h]
\centering
\small
\begin{tabular}{|c|c|c|}
\hline
\textbf{Data Set} & \textbf{Sentences\#} & \textbf{Tokens\#} \\
\hline
Training & 14,987 & 203,621 \\
\hline
Develop & 3,466 & 51,362 \\
\hline
Test & 3,684 & 46,435 \\
\hline
\hline
\multicolumn{2}{|c|}{\# of tag types (IOBES scheme)} & \multicolumn{1}{c|}{17} \\
\hline
\end{tabular}
\caption{NER corpus (CoNLL-2003)}\label{nercorpus}
\end{table}
Performance is measured by the F1 score calculated by the evaluation script of the CoNLL-2003 shared task \footnote{\url{http://www.cnts.ua.ac.be/conll2003/ner/}}.

\subsection{Network Structure}
In all experiments, without specific description, the input layer size is fixed to 100 and output layer size is set as the number of tag types according to the specific tagging task.

\begin{figure}[htb]
	\begin{subfigure}[b]{0.98in}\centering
	\includegraphics[height=0.88in]{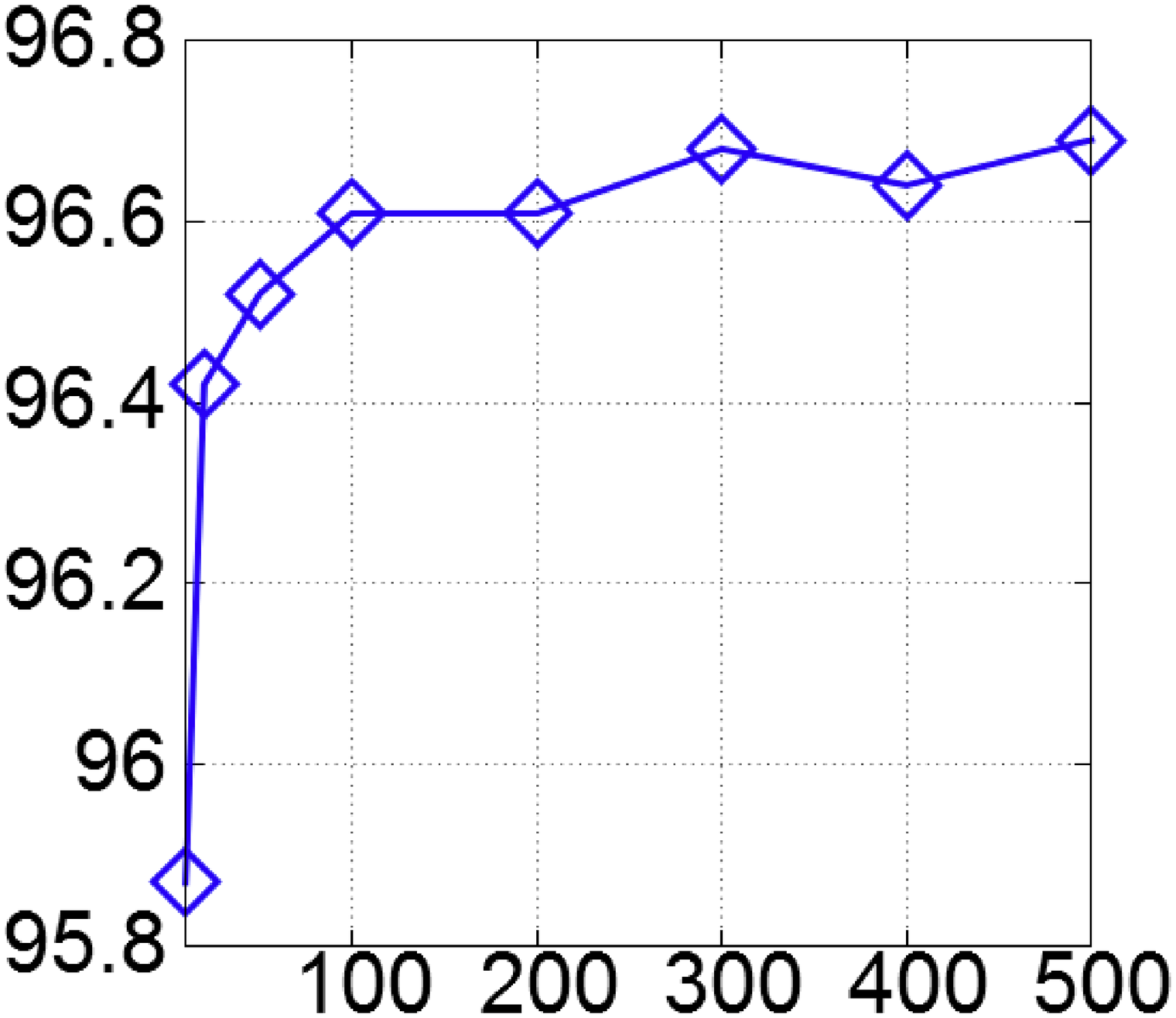}
	\subcaption{POS}
	\end{subfigure}
	\begin{subfigure}[b]{0.98in}\centering
	\includegraphics[height=0.88in]{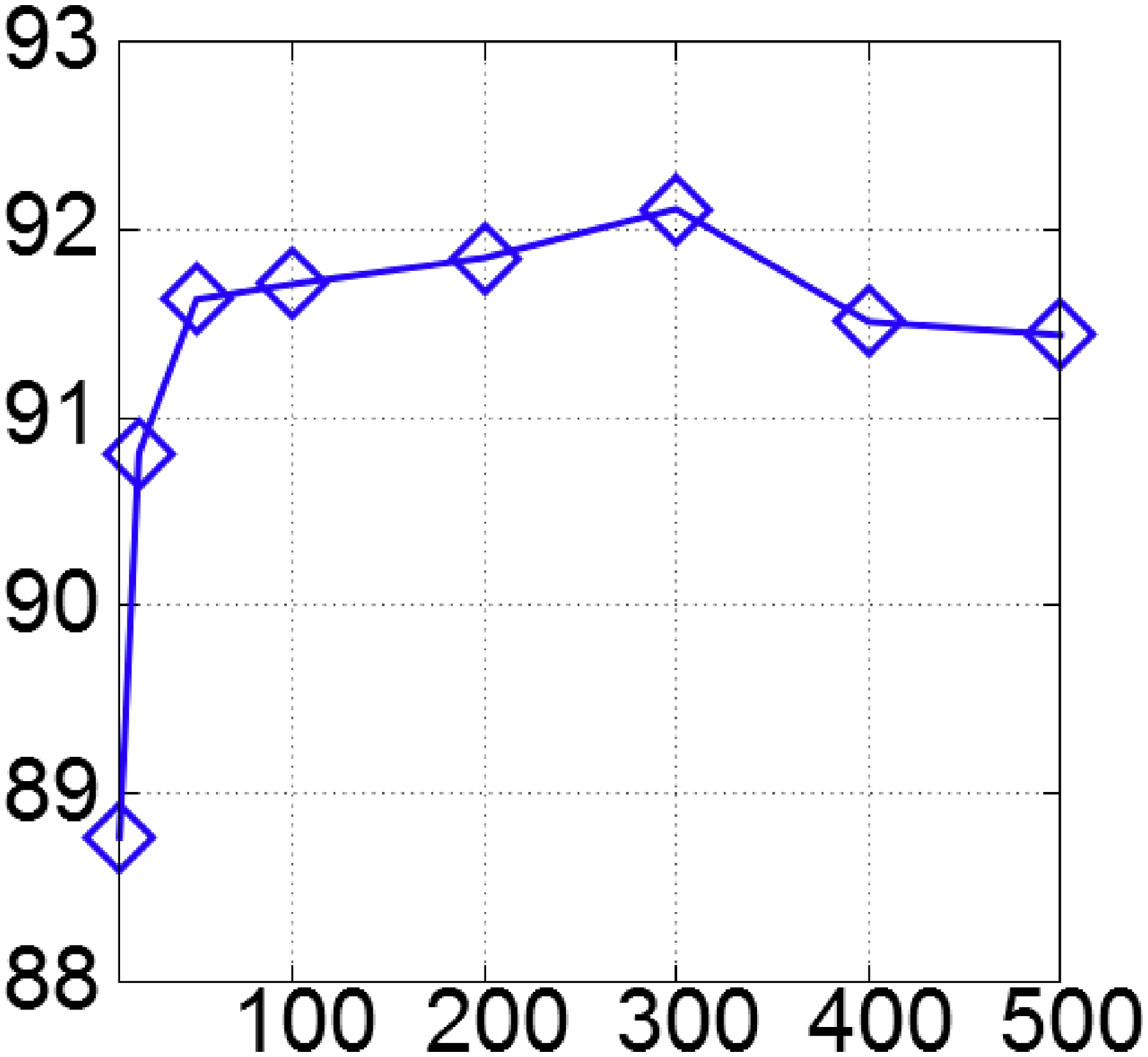}
	\subcaption{CHUNK}
	\end{subfigure}
	\begin{subfigure}[b]{0.98in}\centering
	\includegraphics[height=0.88in]{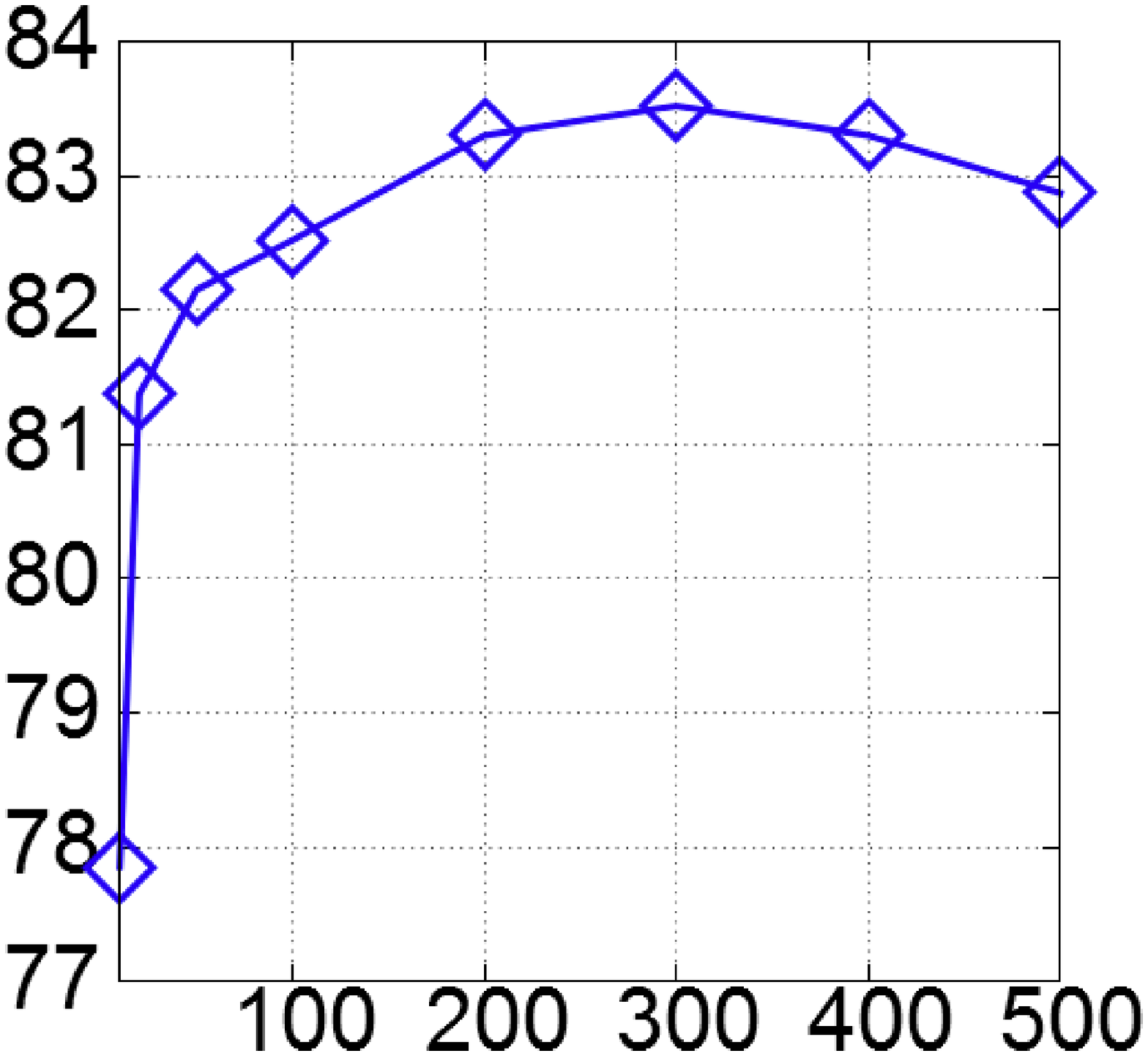}
	\subcaption{NER}
	\end{subfigure}
	\caption{Performance of BLSTM-RNN with different hidden layer sizes. Horizontal axis is the hidden layer size, vertical axis is respectively accuracy for POS, or F1 score for CHUNK and NER.}
	\label{hiddenresult}
\end{figure}

In this experiment, we evaluate different sizes of hidden layer in BLSTM-RNN to pick up the best size for later experiments.
Performances on three tasks are shown in Figure \ref{hiddenresult}.
It shows that hidden layer size has a limited impact on performance when it becomes large enough.
To keep a good trade-off of accuracy, model size and training time, we choose 100 which is the smallest layer size to get a ``reasonable'' performance as the hidden layer size in all the following experiments.

Besides, we also evaluate deep structure which uses multiple BLSTM layers.
This deep BLSTM has been reported achieving significantly better performance than single layer BLSTM in various applications such as speech synthesis \cite{2014_Yuchen_INTERSPEECH_TTSSythesis,2014_Raul_INTERSPEECH_ProsodyContour}, speech recognition \cite{2013_Alex_ICASSP_SpeechRecognition} and handwriting recognition \cite{2008_Alex_NIPS_OfflineHandwriting}.
Table \ref{deeplayercmp} compares the performance of BLSTM-RNNs with one (B) and two (BB) hidden layers.
Size of all hidden layers is set 100.
\begin{table}[h]
\centering
\small
\begin{tabular}{|c|c|c|c|}
\hline
\textbf{Sys} & \textbf{POS(Acc)} & \textbf{CHUNK(F1)} & \textbf{NER(F1)} \\
\hline
B & 96.60 & 91.91 & 82.52 \\
\hline
BB & 96.63 & 91.76 & 82.66 \\
\hline
\end{tabular}
\caption{Comparison of systems with one and two BLSTM hidden layers.  }\label{deeplayercmp}
\end{table}
Using more layers brings a slightly improvement for NER task, while does not show much help for POS and slightly decreases the performance of CHUNK.
A possible explanation is that one BLSTM layer is adequate to learn an effective model for tasks like POS and CHUNK which are relatively simple compared with NER or speech tasks, thus in these cases involving more layers would not provide a further help. 
Meanwhile additional more parameters makes the network harder to converge to a  locally optimal model which leads to a worse performance. 
Based on this observation, in following experiments, BLSTM-RNN is set only one hidden layer.

\subsection{Decoder}
In this experiment, we test the effect of decoder.
\begin{table*}[t]
\centering
\small
\begin{tabular}{|c|c|c|p{7.8em}||c|c|c|}
\hline
\textbf{WE} & \textbf{Dim}& \textbf{Vocab\#} & \textbf{Train Corpus (Toks \#)} & \textbf{POS (Acc)} & \textbf{CHUNK (F1)} & \textbf{NER} \\
\hline
\textbf{\cite{2010_URL_RNNLM}} & 80 &  82K & Broadcast news (400M) & 96.97 & 92.53 & 84.69 \\
\hline
\textbf{\cite{2011_URL_SENNA}} & 50 & 130K & RCV1+Wiki (221M+631M) & 97.02 & 93.76 & 89.34 \\
\hline
\textbf{\cite{2013_URL_Word2Vec}} & 300 & 3M & Google news (10B) & 96.85 & 92.45 & 85.80 \\
\hline
\textbf{\cite{2014_URL_Glove}} & 100 & 1193K & Twitter (27B) & 97.02 & 93.01 & 87.33 \\
\hline
\hline
\textbf{BLSTMWE(10m)} & 100 & 100K &  US news (10M) & 96.61 & 91.91 & 84.66 \\
\hline
\textbf{BLSTMWE(100m)} & 100 & 100K &  US news (100M) & 97.10 & 93.86 & 86.47 \\
\hline
\textbf{BLSTMWE(all)} & 100 & 100K &  US news (536M) & \textbf{97.26} & 94.44 & 88.38 \\
\hline
\hline
\textbf{BLSTMWE(all) + \cite{2011_URL_SENNA}} & 100 & 113K &  US news (536M) & \textbf{97.26} & \textbf{94.59} & \textbf{89.64} \\
\hline
\hline
\textbf{RANDOM} & 100 & 100K & N/A & 96.61 & 91.71 & 82.52 \\
\hline
\end{tabular}
\caption{Comparison different word embeddings.}\label{testwe}
\end{table*}
Performance of BLSTM tagging systems without and with decoder are listed in Table \ref{decoderesult}.
\begin{table}[t]
\centering
\small
\begin{tabular}{|c|c|c|c|}
\hline
\textbf{Sys} & \textbf{POS} & \textbf{CHUNK} & \textbf{NER} \\
\hline
BLSTM & 96.60 & 90.60 & 80.85 \\
\hline
BLSTM+Decoder & 96.60 & 91.71 & 82.52 \\
\hline
\end{tabular}
\caption{Comparison of systems without and with decoder.}\label{decoderesult}
\end{table}
Without using decoder, predicted tag $y'$ is determined by directly selecting the tag with the highest probability among network output $o(w_i)$.
The results show that decoder significantly improves the performance of CHUNK and NER tasks, though shows no help for POS.

For this difference of improvment, we provide a possible explanation.
In tasks like CHUNK and NER which uses IOBES tagging scheme, tags are highly dependent with their previous tags.
For example, \textit{I-X} can only appear behind \textit{B-X}.
CHUNK task has 42 tag types that can combine $42 \times 42 = 1764$ tag bigrams, but only 252 (14.3\%) of them actually appear in training corpus.
In NER task, 78 (27.0\%) of total 289 tag bigrams have occurred more than once.
Decoding in this case filters considerable invalid paths including more than half of candidates and thus improves the performance.
As a contrast, in POS task, tags are directly predicted on each word without using any tagging schemes and thus do not have such strong dependence among tags.
In POS training corpus, 1439 (71.0\%) of total 2025 tag bigrams have occurred more than once.
In this case, the dependences of tags are mainly learnt by BLSTM layer and the help from the decoder is very limited. 

The improvement provided by decoder shows that although BLSTM is considered can adopt contextual information automatically, the model is still far from ideal and a decoder with prior knowledge of tagging scheme is essential for achieving a good performance.

\subsection{Word Embedding}
In this experiment, we evaluate BLSTM-RNN tagging approach using various of word embeddings including those trained by the proposed approach in Section \ref{secwe} as well as four types of published word embeddings.

To train word embeddings with our approach, we use North American news \cite{2008_David_LDC_NorthAmerican} as the unlabeled data.
To construct corpus for training embeddings, the North American news data is first tokenized with the Penn Treebank tokenizer script \footnote{\url{https://www.cis.upenn.edu/~treebank/tokenization.html}}.
Then about 20\% words in normal sentences of the corpus are replaced with randomly selected word.
BLSTM-RNN is trained to judge which word has been replaced as described in Section \ref{secwe}.
To use word embedding, we just initialize the word embedding lookup table ($W_1$) with these already trained embeddings.
For words without corresponding external embeddings, their word embeddings are initialized with uniformly distributed random values, ranging from -0.1 to 0.1.

Table \ref{testwe} lists the basic information of involved word embeddings and performances of BLSTM-RNN tagging approaches using these embeddings,
where RCV1 represents the Reuters Corpus Volume 1 news set.
\textbf{RANDOM} is the word embedding set composed of random values which is the baseline.
\textbf{BSLTMWE(10m)}, \textbf{BSLTMWE(100m)} and \textbf{BSLTMWE(all)} are word embeddings respectively trained by BLSTM-RNN on the first 10 million words, first 100 million words and all 536 million words of North American news corpus.
While \textbf{BSLTMWE(10m)} does not bring about obvious improvement, \textbf{BSLTMWE(100m)} and \textbf{BSLTMWE(all)} significantly improve the performance.
It shows that BLSTM-RNN can benefit from word embeddings trained by our approach and larger training corpus indeed leads to a better performance.
This suggests that the result may be further improved by using even bigger unlabeled dataset.
In our experiment, \textbf{BSLTMWE(all)} can be trained in about one day (23 hrs) on a NVIDIA Tesla M2090 GPU.
The training time increases linearly with the training corpus size.

\begin{table*}[t]
\small
\centering
\begin{subtable}{.3\linewidth}\centering
{
	\begin{tabular}{|c|c|}
	\hline
	\textbf{System} & \textbf{Acc} \\
	\hline
	\textbf{\cite{2012_Liang_NAACLHLT_StructuredPerceptron}} & 97.35 \\
	\hline
	\cite{2003_Toutanova_NAACL_FeatureRich} & 97.24 \\
	\hline
	\hline
	BLSTM & 97.26 \\
	\hline
	\end{tabular}
}
\caption{POS}
\end{subtable}
\begin{subtable}{.36\linewidth}\centering
{
	\begin{tabular}{|c|c|}
	\hline
	\textbf{System} & \textbf{F1} \\
	\hline
	\textbf{\cite{2008_Xu_COLING_ModelingLatent}} & 94.34 \\
	\hline
	\cite{2000_Taku_CONLL_UseOfSupport} & 93.48 \\
	\hline
	\hline
	BLSTM & \textbf{94.59} \\
	\hline
	\end{tabular}
}
\caption{CHUNK}
\end{subtable}
\begin{subtable}{.32\linewidth}\centering
{
	\begin{tabular}{|c|c|}
	\hline
	\textbf{System} & \textbf{F1} \\
	\hline
	\textbf{\cite{2005_RieKubota_JMLR_AFramework}} & 89.31 \\
	\hline
	\cite{2003_Radu_CONLL_NamedEntity} & 88.76 \\
	\hline
	\hline
	BLSTM & \textbf{89.64} \\
	\hline
	\end{tabular}
}
\caption{NER}
\end{subtable}
\caption{Top systems on three tagging tasks. State-of-the-art system is marked with bold type. }\label{syscmp}
\end{table*}

When uses published word embeddings, the input layer size of BLSTM-RNN is set to the dimension of utiembeddings.
All of the four published word embeddings significantly enhance BLSTM-RNN.
It proves that word embedding is a useful feature and is an effective way to make use of big unlabeled data.
Among all mentioned embeddings, \textbf{BLSTMWE(all)} achieves the best performance on POS and CHUNK tasks while slightly falls behind \textbf{\cite{2011_URL_SENNA}} in NER task.
One possible explanation is that \textbf{\cite{2011_URL_SENNA}} is trained on Wikipedia data which contains more named entities, thereby their word embedding contains more useful information for NER task.
 \textbf{BLSTMWE(all)} is trained on a news corpus which is written with more formal grammar, thus it learns better representations for tagging syntactic role. 
Based on this conjecture, it is natural to expect the combination of \textbf{\cite{2011_URL_SENNA}} and \textbf{BLSTMWE(all)} to bring a further improvement.
This idea yields the \textbf{BLSTMWE(all) + \cite{2011_URL_SENNA}}.
This word embeddings are first initialized with \textbf{\cite{2011_URL_SENNA}} and then trained by BLSTM-RNN on North American corpus as \textbf{BLSTMWE(all)}.
This embeddings help BLSTM-RNN obtain the best performance in all three tasks.

\subsection{Comparison with Previous Systems}
 
In this section, we compare our approach with previous state-of-the-art systems on POS, CHUNK and NER tasks.
Table \ref{syscmp} lists the related works of these three tasks.
BLSTM represents the BLSTM-RNN tagging approach using \textbf{BLSTMWE(all) + \cite{2011_URL_SENNA}} word embeddings.

\textbf{POS}: 
\cite{2012_Liang_NAACLHLT_StructuredPerceptron} reports the highest accuracy on WSJ test set (97.35\%).
Besides, \cite{2014_Robert_COLING_FastHigh} (97.34\%) and \cite{2007_Libin_ACL_GuidedLearning} (97.33\%) also reach accuracy above 97.3\%.
These three systems are considered as state-of-the-art systems in POS tagging.
\cite{2003_Toutanova_NAACL_FeatureRich} is one of the most commonly used approaches which is also known as Stanford tagger.
All of these methods utilize rich morphological features proposed in \cite{1996_Adwait_EMNLP_AMaximum} which involves $n$-gram prefix and suffix ($n$ = 1 to 4).
Moreover, \cite{2007_Libin_ACL_GuidedLearning} also involves prefix and suffix of length from 5 to 9.
\cite{2014_Robert_COLING_FastHigh} adds extra elaborately designed features, including flags indicating if word ends with $-ed$ or $-ing$, etc.

\textbf{CHUNK}:
\cite{2000_Taku_CONLL_UseOfSupport} (93.48\%) is the system ranked first in CoNLL-2000 shared task challenge.
Later, \cite{2008_Xu_COLING_ModelingLatent} reports the state-of-the-art F1 score (94.34\%) on CoNLL-2000 task.
Besides, \cite{2003_Fei_NAACL_ShallowParsing} (94.29\%) and \cite{2005_Ryan_EMNLP_FlexibleText} (94.29\%) both report F1 score around 94.3\%.
These systems use features composing of words and POS tags.

\textbf{NER}:
\cite{2003_Radu_CONLL_NamedEntity} (88.76\%) is the top system in CoNLL-2003 shared task challenge.
\cite{2005_RieKubota_JMLR_AFramework} (89.31\%) reports a better F1 score with a semi-supervised approach.
The unlabeled corpus they used is 27M words from Reuters.
Features used in these works include words, POS tags, CHUNK tags, prefixes, suffixes, and a large gazetteer (geographical dictionary).

All of these top systems use rich features and feature sets in different tasks are quite different.
In contrast, our system only uses one set of task-independent features in all tasks and do not require any feature engineering to achieve the state-of-the-art performance in CHUNK and NER tasks.
In POS, our approach also get a competitive performance that is comparable with Stanford POS tagger.
Our results show that feature engineering in conventional methods can be effectively replaced by built-in modeling of BLSTM-RNN.


\section{Related Works}

\cite{2011_Ronan_JMLR_NaturalLanguage} is the most similar work as ours.
It is a unified tagging solution based on neural network which also uses simple task-independent features and word embeddings learnt from unlabeled text.
The main difference is that \cite{2011_Ronan_JMLR_NaturalLanguage} uses feedforward neural network instead of BLSTM-RNN.
A comparison of \cite{2011_Ronan_JMLR_NaturalLanguage} and our approach is listed in Table \ref{cwcmp},
\begin{table}[h]
\centering
\small
\begin{tabular}{|c|c|c|c|}
\hline
\textbf{Sys} & \textbf{POS} & \textbf{CHUNK} & \textbf{NER} \\
\hline
NN & 96.37 & 90.33 & 81.47 \\
\hline
NN+WE & 97.20 & 93.63 & 88.67 \\
\hline
\hline
BLSTM & 96.61 & 91.71 & 82.52 \\
\hline
BLSTM+BLSTMWE & 97.26 & 94.59 & 89.64 \\
\hline
\end{tabular}
\caption{Comparison of \cite{2011_Ronan_JMLR_NaturalLanguage} and our approach. }\label{cwcmp}
\end{table}
where NN is the unified tagging system of \cite{2011_Ronan_JMLR_NaturalLanguage} and NN+WE is that system using word embedding trained by their approach.
Without using word embedding, BLSTM outperforms NN in all three tasks.
It is consistent with the observations in previous works that BLSTM-RNN is a more powerful model for sequential labeling than feedforward network.
With word embeddings, our approach also significantly surpasses NN+WE in all three tasks.

\cite{2008_Suzuki_ACL_SemiSupervised} proposes a semi-supervised approach which incorporates one billion words of unlabeled data during training.
They claim that their model can be applied to various tasks and report the state-of-the-art performance on all these three tasks (97.40 for POS, 95.15 for CHUNK, 89.92 for NER).
However, they also utilize rich feature templates for each task (47 feature templates for POS, 39 templates for CHUNK, 79 templates for NER) which makes their system not so unified.
Their method still requires prior knowledge to design these feature templates which limits its application to tasks that lack of knowledge or tools to extract necessary features.
Besides, they have to train model with quite big unlabeled data for each task which would lead to a time consuming training process.
In contrast, our approach separates the lengthy training for word embedding from the relatively fast training for the supervised tagging task. 
Once our word embeddings are trained, they can be regarded as a kind of linguistic resources like semantic dictionary and be directly used.

\section{Conclusions}
In this paper, we propose a unified tagging solution based on BLSTM-RNN.
This system avoids involving task-specific features, instead it utilizes word embeddings learnt automatically from unlabeled text.
Without reliance on feature engineering or prior knowledge, this approach can be easily applied to various tagging tasks.
Experiments are conducted on three typical tagging tasks: POS tagging, chunking and named entity recognition.
Using simple task-independent input features, our approach gets nearly state-of-the-art results on all these three tasks.
Our results suggest that BLSTM-RNN with word embedding is an effective unified tagging solution and worth further exploration.
\bibliographystyle{acl}
\bibliography{acl2015.mlfornlp}

\end{document}